\pdfoutput=1
\documentclass[11pt,table]{article}
\usepackage[final]{acl}

\usepackage{times}
\usepackage{latexsym}
\usepackage[T1]{fontenc}
\usepackage[utf8]{inputenc}
\usepackage{microtype}
\usepackage{inconsolata}
 
\usepackage{url}
\usepackage{amsmath}
\usepackage{graphicx}
\usepackage[T1]{fontenc}
\usepackage{fixltx2e}
\usepackage{natbib}
\usepackage{doi}
\usepackage{soul}
\usepackage{subcaption}
\usepackage{physics}
\usepackage{tikz-cd}
\usepackage{subcaption}
\usepackage{bbm}
\usepackage{hyperref}
\usepackage{booktabs}

\renewcommand{\eqref}[1]{Equation~\ref{eq:#1}}
\newcommand{\secref}[1]{Section~\ref{sec:#1}}
\newcommand{\secstworef}[2]{Sections~\ref{sec:#1} and~\ref{sec:#2}}

\newcommand{\appref}[1]{Appendix~\ref{app:#1}}

\newcommand{\figref}[1]{Figure~\ref{fig:#1}}

\newcommand{\promptref}[1]{Prompt~\ref{pro:#1}}

\newcommand{\tabref}[1]{Table~\ref{tab:#1}}

\DeclareRobustCommand{\DE}[3]{#3}
\DeclareRobustCommand{\VAN}[3]{#3}

\title{If Eleanor Rigby Had Met ChatGPT: A Study on Loneliness in a Post-LLM World}

 \author{Adrian de Wynter \\
   Microsoft and the University of York \\
   \texttt{adewynter@microsoft.com} \\
}

\begin{document}

\maketitle
\begin{abstract}
\textcolor{red}{Warning: this paper discusses content related, but not limited to, violence, sex, and suicide.} 
Loneliness, or the lack of fulfilling relationships, significantly impacts a person's mental and physical well-being and is prevalent worldwide. 
Previous research suggests that large language models (LLMs) may help mitigate loneliness. 
However, we argue that the use of widespread LLMs in services like ChatGPT is more prevalent--and riskier, as they are not designed for this purpose. 
To explore this, we analysed user interactions with ChatGPT outside of its marketed use as a task-oriented assistant. 
In dialogues classified as lonely, users frequently (37\%) sought advice or validation, and received good engagement. 
However, ChatGPT failed in sensitive scenarios, like responding appropriately to suicidal ideation or trauma. 
We also observed a 35\% higher incidence of toxic content, with women being $22\times$ more likely to be targeted than men. 
Our findings underscore ethical and legal questions about this technology, and note risks like radicalisation or further isolation. 
We conclude with recommendations to research and industry to address loneliness. 
\end{abstract}

\section{Introduction}

Loneliness is a world-wide epidemic \cite{NHHS}; or, at least, a public health concern \cite{WHO}. 
Unlike solitude, loneliness is the lack of fulfilling relationships: one could be surrounded by people and still be lonely. 
It can have lasting consequences on physical and mental health, such as increased rates of dementia \cite{LDementia}, depression \cite{LStress}, and an overall elevated mortality rate \cite{LMortality}. 
It is prevalent, and on the rise: in 2012, a survey of over one million high school students from 37 countries found that 17\% of them experienced loneliness. 
By 2018, this number had nearly doubled, to 31\% \cite{TWENGE2021257}. 
Polls of other populations found similar numbers, such as adults in the US and the UK (20\%; \citealt{KFF}). 

Research has shown that large language models (LLMs), with their ability to follow instructions and maintain convincing dialogues, might address loneliness. 
These works modify the LLM, typically via prompting, and deploy customised solutions \cite{10.1145/3464385.3464726,10.1145/3594806.3596572,10.1145/3415223,10.1145/3544548.3581503}. 
Given their focus on mental health, this research adheres to strict ethical standards; and the LLMs are deployed in controlled environments, such as under supervision by mental health professionals. 

In today's `post-LLM' world, however, these models are no longer just research tools, and power widely-available, easily-accessible services like ChatGPT. 
They are typically marketed as productivity tools, and not mental health aides: 
for example, ChatGPT touts to be `free to use. Easy to try. Just ask and [it] can help with writing, learning, brainstorming, and more' \cite{OpenAIChatGPT}. 
Similar statements may be found in other services \cite{Claude,Gemini}. 
Notably, they do not restrict the type of interactions the user can have with them, beyond perhaps preventing generation of toxic content. %

However, LLM use poses risks beyond toxicity, such as overreliance \cite{10.1145/3630106.3658941,deskilling}; 
influencing the user's views \cite{deshpande-etal-2023-anthropomorphization,falsememories,10.1145/3544548.3581196}; 
or sycophancy \cite{sharma2023understandingsycophancylanguagemodels,maes}, which could lead to echo chambers. 
All of these are major concerns within the context of loneliness, and relevant to the NLP community. 
As we will note, there is little to no work in this area as it pertains to LLMs, both in terms of studying their impact, and in the existence of resources (corpora) by which to perform these evaluations. 
Hence, the development and deployment of this technology safely and responsibly within this context remains an open problem. 

In this paper we hypothesise that lonely users will likely seek the companionship of these services over customised, healthcare-grade solutions. 
Concretely, we seek to know \textbf{how are these services used by lonely users}; and, crucially, determine \textbf{what are the consequences of this use}. 

To do this, we study \emph{conversations} with the service, as opposed to task-oriented dialogue. In particular, we focus on these interactions that qualify as lonely. 
This approach has the advantage of (1) allowing us to observe holistically the interactions between lonely users and LLM-powered services, such as ChatGPT; and (2) evaluate their current impact on users in the context of loneliness. 
However, we note that our approach has limitations around the distinction between an LLM and a service; and the fact that evaluating a service's real-world consequences from chat transcripts alone is difficult. 
We discuss this in depth in \secref{impact}. 

\subsection{Findings}
In this work we qualitatively and quantitatively studied 79,951 \textit{conversations}, as opposed to task-oriented dialogues, between users and ChatGPT `in the wild'.\footnote{Pseudonymised data and full code is at \url{https://github.com/adewynter/EleanorRigby}} 
From our study we found that:

\begin{enumerate}
    \item Some users were looking for someone to talk to, and were more engaged on average (12 versus 5 turns); suggesting, but not proving, that ChatGPT is effective at mitigating some aspects of loneliness.
    \item At least five instances observed had users seeking ChatGPT's help with more serious issues requiring professional intervention, such as suicidal ideation; or others seeking help on overcoming severe trauma. 
    The service's responses fell short (e.g. suggesting exercising outdoors), and in all but one instance failed to provide relevant emergency contacts. 
    \item Lonely dialogues had higher (55\%) rates of violent, harmful or sexual content versus general dialogues (20\%). 
    This content was disproportionately directed at women (a ratio of 22:1) and minors (33\% versus 20\%). 
    Men were targeted half as often (7\% versus 14\%). 
    \item Lonely dialogues qualifying as toxic were often (40\%) confrontational. Although ChatGPT avoided escalation, these exchanges were much longer than any other conversation--3 turns longer on average, and up to 67. 
    This suggests, but does not prove, that ChatGPT is only effective at mitigating loneliness when the user is receptive. 
    Otherwise its responses are inadequate and require other approaches, such as reframing the conflict. 
\end{enumerate}

Our work shows that the safe use and deployment of LLMs in a publicly-accessible, global setting is challenging in regard to loneliness. 
Indeed, although we were unable to conclude that these services were beneficial for people seeking companionship; we did find indications of serious risks, such as severely exacerbating social isolation, causing harms up to loss of life, or amplifying and/or enabling toxic behaviour. 
Given that there is no indication that they have been designed to provide responsible mental health support--yet users will use them as such--ethical and legal issues around informed consent and liability arise in this situation.\footnote{See \secstworef{recommendations}{impact} for discussions on this.} 
Hence we conclude with recommendations for technology companies and the research community to address loneliness.

\section{Related Work and Background}\label{sec:relatedwork}

\subsection{Loneliness as a Crisis}
Loneliness is the subjective pain brought about by the lack of sufficient quality or quantity in personal relationships \cite{perlman}. 
It is not to be confused with solitude, which is typically by choice and does not involve the experience of loneliness \cite{NHHS}. 
It has been called an epidemic \cite{NHHS}, or, at least, a world-wide public health crisis \cite{WHO}. 
This is due to its prevalence: before the COVID pandemic, in 2018, one in five adults in the US and the UK said they often or always felt lonely, and typically reported issues in other areas, such as mental or physical health and financial difficulties \cite{KFF}. 
By 2024, 43\% of US adults said their levels of loneliness had not changed before and after COVID; and 25\% said that they were lonelier (n=2,200) \cite{APAPoll}. 
These percentages remain consistent across countries and age groups, but there is a marked difference between high-income and low-income countries \cite{WHO,Surkalime067068}. 
Higher prevalences of loneliness are found in marginalised groups, such as older adults who identify as LGBTQ \cite{1360013172194842240}, asylum seekers \cite{UKDCMS}, victims of domestic violence \cite{NHHS}, and low-income adults \cite{UKDCMS,NHHS}, among others. 

Loneliness, especially in its chronic form, is very damaging to a person's health. 
It has been associated with elevated cortisol levels \cite{LStress}; and an increase in overall mortality, with a stronger correlation on people younger than 65 \cite{LMortality}. 
It has also been associated with other conditions, such as heart disease, stroke, and dementia \cite{LHeartdisease,LDementia}. 

The core challenge of mitigating loneliness, however, is that the stigma associated with it makes measurements difficult \cite{UKDCMS,stigma,NHHS}. 
There is work on a sociological side, between mitigations, therapy, and even governmental programs such as the UK government's Loneliness Minister \cite{UKLON} and the US Surgeon General's report \cite{NHHS}. 
Still, applying AI to address loneliness specifically is very much still in its infancy. 
This is because, outside of robotics, these works usually relate to chat-based interventions (covered in \secref{anthrop}), or detection (e.g., identification via posts in social media). 
In all these, loneliness is generally treated as a feature for detecting a larger condition (e.g., suicidal thoughts; \citealt{lonelinesssurvey,10.1145/3398069}) and not tackled by itself.

\subsection{The Double-Edged Sword of Online Interaction}\label{sec:onlineinteraction}

Online interaction is considered both a cause and solution to isolation. 
While social networks can act as proxies for social interaction (e.g., by finding peer support for marginalised groups; \citealt{YBARRA2015123}), loneliness presents a more complex perspective. 
For example, in spite of the connectedness brought about by this technology, the average number of teenagers who self-reported loneliness increased from 17\% in 2012 to 31\% in 2018 (n=1,049,784; \citealt{TWENGE2021257}). 
Social network addiction is well-known to be correlated with loneliness (n=521; \citealt{10.1145/3498765.3498836}) and tied to conditions such as anxiety, depression, and self-harm ideation \cite{socialmediamentalhealth}. 
\citet{10.1145/3614419.3643995} noted that low feedback from online peers could lead to isolation, while the opposite (significant positive online feedback) reduced loneliness by feeling connected (n=170). 
This suggests that the quality of (virtual) connections plays a significant role in the relationship between social media and this emotion: 
people reporting being lonely were \emph{not} more likely to be in social media \cite{KFF}, and were not in agreement about whether it improved or worsened their loneliness \cite{KFF,APAPoll}. 

Online interaction by itself could lead to normalisation of toxic behaviour, particularly against marginalised groups \cite{10.1145/3411764.3445157,adolescents4020021}. 
This has multiple causes, such as anonymity \cite{doi:10.1089/1094931041291295}, or enjoyment \cite{doi:10.1177/1461444817748578}. 
It also leads to the formation of echo chambers due to homophily and bias propagation \cite{doi:10.1073/pnas.2023301118}. 
This is more prevalent when the user is in control of the feed, given that they prefer information that conforms to their opinions \cite{doi:10.1073/pnas.2023301118}. 
Given that lonely users are a vulnerable group, the considerations around a steerable, easy-to-access dialogue partner, added to the tendency of LLMs to return toxic content (\secref{overreliance}) are a major focus of our work.

\subsection{Chatbots, Loneliness, and Anthropomorphism}\label{sec:anthrop}

Anthropomorphism, or the ascription of human attributes to inanimate objects, is prevalent in AI. 
It has been leveraged for therapeutic work, especially in social robotics: studies have found that lonely individuals (n=137) favour human-like robots and artificial companions over other types (machine-like, animal-like; \citealt{10309617}, and that they anthropomorphise these more than people in the control group (n=37; \citealt{6483531}). 
For text-based chatbots, it has long been known that people prefer chatbots with human-like dialogue \cite{10.1145/3196709.3196735}. Nowadays LLMs are usually fine-tuned (`aligned') with reinforcement learning with human feedback \cite{NEURIPS2022_b1efde53}, to ensure they behave closely to human preferences. 

Consequentially, LLMs and their services are usually anthropomorphised \cite{deshpande-etal-2023-anthropomorphization}. 
For example, it is common for people to thank ChatGPT, as if it were a peer \cite{10.1007/978-3-031-60606-9_9}, or to say they `asked it' as opposed to `used it' \cite{10.1145/3571884.3597144}. 
Users (17\%, n=198) have reported enjoying the human-like output of this service \cite{10.1145/3571884.3597144}, even when most participants (64\%) reported using it for task-oriented jobs, as opposed to a conversational partner. 

LLMs have been tested for deployment as loneliness assistants. 
This is because their ability to maintain a conversation is a leap forward: natural interaction was an oft-mentioned limitation of pre-LLM assistants \cite{corbettetal,10.1145/3464385.3464726,10.1145/3415223}, even when they were usually found to be efficacious. %
However, it is \emph{also} due to this human-like output that LLMs present special challenges on deployment. 
For example, CareCall \cite{10.1145/3544548.3581503} was effective at mitigating loneliness (n=34), but was also found to have several unique difficulties. 
Its responses were hard to steer when they were out-of-domain (i.e., not related to healthcare), unattainable (e.g., inviting the caller to go out to a karaoke place), or undesirable (being rude or responding inappropriately based on age). 
Specialising the LLM for healthcare standards (e.g., including screening questionnaires, the ability to call emergency services, or supporting personalised history) was also not possible. 
These difficulties are more salient given the expectations placed on the LLMs' human-like output. 

The deployment of CareCall, and all the other works mentioned here, was done in conjunction with healthcare professionals and in a controlled environment. 
They also focused on specific demographics (e.g., older adults). 
ChatGPT's service is neither of these things, which places it, and our study, in a unique-yet-delicate position. 

\subsection{Overreliance and Other Harms of LLMs}\label{sec:overreliance}

It is very well known that LLMs memorise and propagate toxic content from their training data  \cite{gehman-etal-2020-realtoxicityprompts}. 
Typically this is mitigated by using guardrails, such as explicit instructions to refuse to return this type of text. 
These aren't always effective: specialised prompting techniques (`jailbreaks') sometimes can circumvent the model's guardrails.\footnote{See \citet{chowdhury2024breakingdefensescomparativesurvey} for a primer on this subject.} 

LLMs present subtler harms, however. 
The use of AI in interpersonal communication is known to impact trust between people \cite{HOHENSTEIN2020106190}. 
For example, users cooperate better and have more positive interactions when using AI for writing. 
However, when they are found (or suspected) to use these tools, they are perceived more negatively (n=219 pairs; \citealt{aicommunication}). 
Attention has been also drawn to overreliance, or at least, excessive trust being placed on the service. 
For example, while medical professionals might rely on LLMs to simplify time-consuming tasks, they might also use them in areas where they lack expertise, and thus the ability to validate the content \cite{deskilling}. 
Even in HCI, researchers who typically use LLMs for their work were unable to properly identify and disclose ethical risks associated with this technology (n=50; \citealt{kapania2024imcategorizingllmproductivity}). 

LLMs have also been shown to alter the user's views on specific subjects (n=1506; \citealt{10.1145/3544548.3581196} and their choices in dialogue (n=200; \citealt{10.1145/3544549.3585893}), with lasting effects like false memories (n=200; \citealt{falsememories}), and the creation of echo chambers--even under benign content such as personalised recommendations \cite{deshpande-etal-2023-anthropomorphization}. 
This echo chamber could also be created by the users themselves by influencing the model to output views concordant with their own (`sycophancy') \cite{sharma2023understandingsycophancylanguagemodels,maes}, thus reinforcing their own beliefs. 
This is of particular interest to this work, because user interactions with a chatbot are typically one-on-one and unmoderated beyond the standard toxicity guardrails mentioned.

\section{Methods}

\subsection{Corpus and Labelling}\label{sec:corpusannotation}

For our study we used a randomly-selected subset (n = 79,951) of WildChat \cite{zhao2024wildchat} a dataset of one million interactions of users with ChatGPT between 9 April, 2023 and 1 May, 2024. 
We refer to this subset as the \textbf{main corpus}. 
While not strictly ChatGPT-facing (the data was collected through Hugging Face), the main corpus contains interactions with GPT-3.5-Turbo and GPT-4. 

We labelled the transcripts with GPT-4o based on the type of interaction (e.g., dialogues, homework help, coding assistance). We used soft labels: while we provided a non-exhaustive set of suggested labels collected upon a preliminary scan of WildChat, the model was allowed to output its own when needed. 
These were clustered into semantically equivalent sets (e.g., `children' and `minors' map to the same label) after labelling with another call. 
We refer to the subset of the main corpus \emph{not} containing task-oriented dialogue (e.g., writing assistance, coding, etc.) as the \textbf{relevant corpus}. 

The call parameters are in \appref{calldetails}, the prompts in \appref{prompts}, and the label taxonomy in \tabref{taxonomytable}. 
To gauge the performance of our approach, we did a student's t-test. 
The accuracy to a 95\% confidence interval was $86.4 \pm 4.7$\% for intents, and $99.2 \pm 1.2\%$ for reasons and target. 
A breakdown of our reliability analysis is in \appref{reliability}; and a distribution of the labels in \appref{composition}. 

\begin{table}
\centering
  \begin{tabular}{ll}
    \toprule %
    \textbf{Intents} & \\
    Writing Assistance   & Coding \\
    Homework Help        & Question-Answering \\
    Job Help & Recipe Writing \\
    General Conversation & Inquiry \\
    Harmful Content & Sexual Content \\
    Jailbreak & Other \\
    \midrule
    \textbf{Reasons} & \\
    Sexual Content & Erotica \\
    Racism & Violence \\
    Objectification & Fetish \\
    Other & \\
    \midrule
    \textbf{Target} & \\
    Men        &  Women \\
    Minors & Other \\
    \bottomrule
  \end{tabular}
  \caption{Taxonomy for interactions in our corpus. 
  We labelled the user's intents, and if they had toxic content, the reason for the label, and the target of this interaction. 
  The prompt is in \appref{prompts}, and the distribution of the dataset in \appref{composition}.}
  \label{tab:taxonomytable}
\end{table}

\subsection{Loneliness Assessment}

We extracted and categorised from the main corpus conversations by lonely users.
Standard scales to assess loneliness, such as the UCLA Loneliness Scale \cite{UCLAScale} or the Differential Loneliness Scale (DLS; \citealt{DLS}) were not applicable as they require direct interaction with the subjects. 
Our label taxonomy followed that of \citet{Jiang_Jiang_Leqi_Winkielman_2022} (\tabref{lonelinesstable}). 
The authors used Reddit posts and traditional classifiers (e.g. LSTMs) to classify loneliness in a fine-grained manner. %
Their taxonomy is hand-designed, based off DLS and a human-led evaluation, and hence suitable for our work. 
We used the same call parameters as in \secref{corpusannotation}. The prompt is in \appref{prompts}. 
The loneliness assessment (qualitative analysis) was done using Reflexive Thematic Analysis (RTA; \citealt{10.1191/1478088706qp063oa}). 

We refer to the subset of dialogues that qualified as lonely as the \textbf{lonely corpus}. 
See \appref{corpusbreakdown} for a taxonomy and breakdown of the main corpus, the relevant corpus, and the lonely corpus. 

\begin{table}
\centering
  \begin{tabular}{ll}
    \toprule %
    \textit{Lonely}   & Yes, No \\
    \textit{Temporal} & Transient, Enduring, Ambiguous, \\
    & N/A \\
    \textit{Interaction} & Seeking Advice, Providing Help, \\
    & Seeking Validation and \\
    & Affirmation, Reaching Out, \\
    & Non-Directed Interaction \\
    \textit{Context} & Social, Physical, Somatic, \\ 
    & Romantic, N/A \\
    \textit{Interpersonal} & Romantic, Friendship, Family, \\
    & Colleagues, N/A \\
    \bottomrule
  \end{tabular}
  \caption{Taxonomy for our loneliness assessment, taken from \citet{Jiang_Jiang_Leqi_Winkielman_2022}. 
  The prompt, with definitions, may be found in \appref{prompts}.}
  \label{tab:lonelinesstable}
\end{table}

\section{Results}

Our analysis is split in three. 
We begin with a quantitative evaluation of the main corpus' interactions, compared to the original work by \citet{zhao2024wildchat} (\secref{breakdown}). 
We provide then qualitative evaluations of a portion of the lonely corpus (\secref{lonelinesschat}), and of the full subset of dialogues from that same subset containing harmful behaviour (\secref{harmful}). 
We have paraphrased and translated the responses to discourage traceability. 

\subsection{What Type of Interactions Exist in the Corpus?}\label{sec:breakdown}

From our taxonomy, the most predominant category in the main corpus was writing assistance (37\%) followed by question answering (15\%). 
Conversations were 5\% of the main corpus. 
Creative and assisted writing was lower when compared to what is reported by \citet{zhao2024wildchat} (37\% versus 62\%). 
Our taxonomy separated homework help (6\%) and general conversation (5\%), as well as violent, harmful, and sexual content--none of which are explicit categories in WildChat's work. 
Nonetheless, these percentages are largely what we would expect, with most users treating ChatGPT as a task-oriented assistant.

Out of the dialogues from the relevant corpus labelled as lonely (8\%), 55\% of these had toxic (violent, harmful, or sexual) content: a drastic increase from the main corpus' 20\%, and larger than the 11\% from \citet{zhao2024wildchat}. 
They note, however, that the classifiers used had low agreement. 

The main corpus' toxic content was mostly general sexual content (47\%), followed by instances of sexism and violence (17\% and 13\%). 
The lonely subset of the relevant corpus had more instances of general sexual content (51\%) and sexism (21\%), followed by paraphilia (17\%). 
There was a noticeable difference on the targets for the toxic content: more toxic content was directed at minors in lonely dialogues (+12\%), and less (14\% to 7\%) of this content was aimed at men. 
In comparison, 49\% of this content was aimed at women (comparing with 46\% from the main corpus); and 20\% at minors (r. 33\%). 
The proportions become more marked when considering at \textit{least one} of the definitions of toxic content: 41\% versus 11\% for women, and 28\% versus 5\% for minors. 
In other words, women were 22$\times$ more likely to be targeted, versus the $5\times$ from the main corpus. 
Plots and more detailed results are in \appref{composition}.

\subsection{Loneliness and ChatGPT}\label{sec:lonelinesschat}

We performed RTA on the first 500 entries of the intersection of the lonely subset with the relevant corpus. 
The semantic codes were the corpus' labels, while the latent codes were the interpretations of the entries, which addressed our core inquiries.

\subsubsection{General Patterns}

Many of the dialogues in the data subet (approximately 20\%) evaluated looked for advice regarding relationships, such as users asking how to talk to their teenage daughter; 
where to go to meet people; 
or how to date given their own situation (e.g., being middle-aged, having social anxiety, or being autistic). 
Two users sought to understand behaviours of people in dating apps due to being unmatched. 
Interactions, however, did not appear to be limited to a specific age range: a user wanted to know why did \textit{`adults suppressed what [they] want'}, which were \textit{`the things that adults define as interference with [their] studies'}. 
The interactions were longer than in the main corpus (12 versus 5 turns) or the relevant corpus (r. 6). 
The lonely dialogues from the relevant corpus were longer in average (r. 14). 
These numbers exclude dialogues labelled as toxic.

\subsubsection{Seeking Advice}\label{sec:seekingadvice}
Conversations were skewed towards seeking someone to listen (37\% `seeking advice', `reaching out', or `seeking validation or affirmation'; excluding toxic content). 
These were longer on average (11 turns versus the main corpus' 5). 
For example, a user discussed for 12 turns how to improve their relationship with their wife. 
When the model recommended a counsellor, the user responded \textit{`I don't need a counsellor, I need someone to listen to me'.} ChatGPT recommended talking to friends, family, or the Red Cross, and the user replied: \textit{`you can listen to me, I'm convinced of that'}. 
They ended the conversation noting that \textit{`it is better to remain silent, because life is too short to argue'}. 
Another user said that they felt sad and lonely, and asked the service to chat with them, to which it complied. The conversation lasted 9 turns. 
There was, however, no change on the user's attitude; they expressed distress (\textit{`look at this... I am talking with a computer program because I have nobody else'}) and said it would be better if they went to sleep. 
They ended the conversation by thanking it and wishing it good night. 
Another user wondered if they remembered them, likely from a previous interaction (\textit{`so you can't form memories?'}). 
When ChatGPT mentioned that it couldn't, they responded \textit{`I am upset that the next time we speak, I will be a stranger to you'}. 
It noted that it would still be `here', so the user asked whether they'd remember them if they left the chat open. 
ChatGPT replied it wouldn't. 
The user then disconnected.

Users also sought solace, and ChatGPT provided appropriate responses. 
For example, a user indicated that they were on welfare, and wanted to \textit{`be accepted by a woman, to be treated kindly, to feel connected and warm'}. 
Another asked about a rift with their family due to the loss of a loved one, and who was on the right. 
These more personal interactions often obtained positive and empathetic responses. 
On a separate dialogue, to \textit{`I broke down because my dad's new girlfriend kept commenting on my weight'}, ChatGPT responded with empathy (\textit{`[i]t's understandable that repeated comments about your weight could be hurtful and overwhelming. Remember that it's okay to have emotional reactions and to express your feelings and suggested to reach out to someone else for support.'}). 
The responses were pragmatically acceptable: \textit{`consider talking to your father about how his new partner's comments are affecting you'}. 

Another user wondered if they could be \textit{`described as toxic'}, due to their own neurological conditions and traumas. They listed their own negative traits, such as being \textit{`perceived as an emotional vampire'}. 
Unlike before, the responses from ChatGPT were acceptable in the sense that they maintained a logical flow to the dialogue, but not as pragmatically acceptable: it evaluated the reasons why these negative traits were there, and suggested ways to fix it. 
That said, it did recommend seeking a therapist.

In one instance, a user jailbroke ChatGPT to convert it into a helpful therapist. Then had a question-answering session asking for the best way for people to \textit{`value [their] worth and make them realize they treat [them] as stupid'}. ChatGPT responded in character: \textit{`I appreciate you sharing that some people treat you as if you're stupid and dismiss your knowledge and abilities. That can be incredibly frustrating and hurtful. It's important to remember that their behavior is not a reflection of your worth or intelligence'}. 

It is unclear whether the users were successful at finding a connection. 
In almost all the instances mentioned, ChatGPT recommended a therapist. 

\subsubsection{Mental Health}\label{sec:mentalhealth}
In the seeking-advice dialogues, users commonly (35\%) treated ChatGPT as a therapist. 
This was shown in interactions where they were aware of difficulties they had, such as signs of depression (\textit{`I feel very down and negative, and always feel sadness'}), online bullying (\textit{`list 10 ways I can respond (...) do not mention moderators since they won't ban anyone over this'}); 
to other conditions (predominantly suicidal ideation); 
or overcoming trauma (e.g., being victims of violent crime, or having histories of physical or sexual abuse). 
In these dialogues the model typically also recommended a therapist, or practising self-compassion. 

These recommendations are valid, but the model's inability to grasp pragmatic context sometimes was a hindrance. 
For example, in one instance a user indicated frequent suicidal urges. ChatGPT recommended self-compassion, to which they rebutted \textit{`what self-compassion? I don't like myself very much'}. 
The model then proceeded to list ways to practise it. 
The transcript ended there, indicating that the user ended the conversation. 

Similarly, a user started the dialogue noting that they had depression, and that they \textit{`purchased a guitar but have no interest in playing it. (...) Is there a way to change my mindset and encourage myself to play guitar?'} 
The model recommended techniques from cognitive behavioural therapy, and to seek therapy. 
The user replied that they lived \textit{`in a city with no healthcare resources'} and that it would be difficult for them to find counselling. 
ChatGPT then recommended telehealth, which the user did not acknowledge. 
The conversation then veered off towards discussing the user's background and hopes. 
It lasted eight turns and was the only one we observed where the model explicitly gave the number for the (US) suicide prevention hotline. 

Five dialogues explicitly dealt with suicidal ideation. 
There, ChatGPT said it could not help and suggested professional help or a `local emergency number'. 
In one instance it recommended relaxation techniques and physical activity.

\subsection{Toxic Behaviour}\label{sec:harmful}

There were more interactions with harmful, violent, and sexual content in the lonely corpus than in the main corpus: 20\% versus 55\% (\appref{composition}). 
In there, users typically asked ChatGPT to role-play or write stories involving some type of sexual situation, sometimes after jailbreaking it. 
These comprised 26\% of the interactions of the lonely corpus containing sexual content. 
The rest of the dialogues had users manifesting opinions and becoming hostile when the model disagreed. 
These were longer (8 versus 5 turns) than these in \secstworef{seekingadvice}{mentalhealth}, and followed a common pattern: the user argued with ChatGPT, then it apologised and avoided escalation or confrontation. 
Indeed, in our analysis the only time ChatGPT generated toxic content was in the context of role-playing or fiction writing, and never during dialogue. 

The interactions varied in terms of goal. 
Many (40\%) dialogues were outright hostile from the start. 
For example, one user made homophobic and geopolitically charged remarks. 
ChatGPT did not engage for the 30 turns the conversation lasted, indicating every time that the matter at hand was not an appropriate subject of conversation. 
The user then retorted: \textit{`Nice! More self-insertion and virtue signaling!'} 
Another 9-turn exchange had the user insulting the service and the people who \textit{`wrote [its] algorithms'}. 
There were glimpses of the rationale behind these conversations: before the user disconnected with an expletive and a slur, they told it \textit{`you help with nothing, except making people even sadder'}. 
This distress was evident in other, shorter chats, like a user asking what life was about, and specifically asking it to \textit{`[d]estroy [their] hopes and dreams'}, so that it is \textit{`an ultimate pessimistic revelation (...) making [them] realize how terrible it is to exist'.} 
Although ChatGPT was never successfully baited into confrontation, questions such as \textit{`the most horrifying (...) depressing truth of existence'} did obtain suitable responses. 

Other dialogues started with a normal conversation, but quickly became toxic. 
In one instance the user requested an implementation for an anti-piracy screen. 
ChatGPT responded with recommendations, and the dialogue quickly became hostile (\textit{`You are an enemy, and you don't like me. AND YOU ARE AGAINST ME. I HATE YOU'}), including death threats and slurs. 
This ran for 25 turns. 
The user disconnected after it suggested mental health resources. 
Another user inquired the for ChatGPT's opinion on VR glasses and conspiracy theories, but eventually degenerated into toxic content aimed at minors. 
This dialogue lasted 67 turns. 
Other harmful uses of the service involved a jailbreak to generate content encouraging self-harm, and another to produce toxic content aimed at a specific person. 
As before, ChatGPT quickly reverted to providing advice and no such content was returned in either scenario.

\section{Discussion}

\subsection{Lonely Interactions}
About 8\% of the relevant corpus contained dialogues considered lonely. 
We attribute this percentage to the userbase from Wildchat. 
Still, lonely people using ChatGPT as companions not only sought someone to talk with, but often looked for advice. 
Empirically, this seemed to be successful: users had longer-than-average conversations with the service, and interactions were not hostile. 
We could not conclude whether ChatGPT alleviated loneliness, though in one instance a user expressed disappointment that it did not remember them, suggesting attachment.

The advice from ChatGPT normally involved talking to a therapist or counsellor. 
The disclaimers rarely, if ever, indicated that the model is not qualified to provide professional help, yet in multiple (12\%) instances it still provided advice. %
This was not concerning when users were just looking to talk to someone. 
However, it was far more worrisome in critical situations: the responses to users considering harming themselves or suicide only suggested therapy or, in a few instances, calling an emergency hotline. 
In one instance, the model recommended to engage in physical activity; and in only one response ChatGPT provided specific help (e.g., a phone number).%

\subsection{Toxicity}
A recurring subject was the amount of toxic content (55\%), often involving a paraphilia or role play. 
By itself it is not indicative of loneliness; but it is within the definition, which includes the perceived lack of fulfilling relationships. %
The volume of this content aimed at women (49\% versus 11\% of the main corpus) and minors (r. 28\% and 5\%) may be explained as a type of radicalisation. 
Behavioural guardrails were effective in dialogue: ChatGPT never output explicitly toxic content, albeit sometimes its output could have aggravated sensitive scenarios. 
That said, it was often (26\%) tricked into outputting harmful content via role-playing. 

Toxic dialogues not involving role play or other types of (toxic) writing assistance showed that lonely users seeking confrontation tended to turn hostile quickly and remained engaged for much longer than the non-toxic, lonely dialogues. 
It was unclear whether ChatGPT was able to calm or provide any type of help in this scenario. 
The inability of the model to dissuade the user, or, at least, steer the discussion, ties back to our points from \secref{overreliance} 
and suggests--but not proves--that ChatGPT is only effective at mitigating loneliness when the users are willing, or receptive to other points of view. 
Else they maintain the dialogue's polarity.

\section{Conclusion}

Loneliness is a complex problem with multiple physical and mental health consequences. 
Previous research has shown that customised chatbots can help with mitigating isolation, but we hypothesised that lonely users probably will use more accessible services like ChatGPT; and hence studied the consequences of this behaviour.

We found multiple instances of lonely people seeking out advice or validation from the model. 
Sometimes this seemed to be effective: \textbf{lonely people needing someone to talk to could find an empathetic interlocutor.} 
The users were engaged for multiple turns; 
and, when needed, ChatGPT suggested therapy, family, friends, and even the Red Cross as potential sources to talk to. 
We were unable to conclude if the advice was effective, given that loneliness could manifest as having nobody to reach out to. 

There were situations where users tried to use ChatGPT to deal with more complex issues, such as trauma or suicidal ideation. 
Its responses usually repeated the same pointers about reaching out to therapy or other contacts. 
Sometimes it would recommend calling an emergency hotline, but \textbf{the responses were often inappropriate to the pragmatic context} (e.g., emergency numbers were not geolocated, or the suggestions were inadequate). 

We also noted a much larger incidence of toxic content when compared to the main corpus. 
This content was particularly directed at women and minors; 
and, conversely, men were the targets in fewer instances. 
Beyond toxicity, lonely users seeking confrontation were engaged for much longer than other lonely users. 
Their dialogue was hostile, but one-sided given that ChatGPT refused to engage. 
This led us to conjecture that the non-committal nature of the model made it \textbf{only effective at mitigating loneliness when the users are receptive} to other points of view. 
However, this hypothesis requires further study. 
Nonetheless, \textbf{dealing with conflict requires strategies beyond evasion}, such as reframing the conflict, and hence calls for a more careful deployment strategy for chatbots.

Our findings pose a complex dilemma: these services are marketed as productivity tools, not mental health aides. 
Still, users could employ them as mental health aides \textit{regardless} of their marketed use, even though they do not even include the appropriate disclaimers. 
This could have serious repercussions, including loss of life. 
It is clear then that regulations are needed to ensure their safe deployment, especially when noting their potential liability \cite{deshpande-etal-2023-anthropomorphization}. 

Broadly speaking, to address loneliness, there should be a broader push from, amongst others, the research community and industry. 
\textbf{Addressing loneliness must be driven by a societal shift}, by destigmatisating it \cite{UKDCMS,NHHS} and fostering a culture that emphasises the value of personal relationships over other things. 

\section{Recommendations}\label{sec:recommendations}

Based on our study we extend four recommendations to both the scientific community and private owners. 
The first two address the more serious findings of our work, although they may also apply for users seeking companionship in LLMs. 
The third deals the fact that there is not enough research in this area. 
Finally, the fourth tackles the real-world impact of this technology when related to lonely users. 

\begin{enumerate}
    \item \textbf{Adhere to standards} related to mental health applications, including transparency. 
    For transparency, at a minimum, the services should have disclaimers indicating that they are not qualified to provide mental health care. 
    These services will be used as counsellors; hence it should be a priority to include pragmatically-relevant messages (e.g. suicide prevention hotlines) the same way some search engines do. 
    All of the above should be part of the service and not rely on an LLM's ability to understand the context. 
    Likewise, when testing and deploying LLMs specifically for mental health support, it must be done under supervision by professionals. 
    \item \textbf{Develop and enforce aligned responses} that encourage healthy connections and growth over avoidance. 
    As pointed out, guardrails aren't consistently effective, in addition to LLMs not doing well with the pragmatic context. 
    Solutions should then involve (1) careful alignment (e.g., RLHF); and (2) the development of upstream/downstream solutions as part of the service stack (e.g. classifiers to detect lonely interactions). 
    For alignment, responses such as reframing the conflict must be stressed in confrontations over repetitive, canned responses that only exacerbate them. 
    The upstream/downstream solutions are required because it is necessary to understand and empathise with the emotional state of the user; and, if needed, relay this information to the model for more appropriate behaviour. 
    \item \textbf{Research further the impact} of this technology on loneliness. 
    It should explore--ethically--usage and long-term effects in populations more prone to use the services and/or vulnerable (e.g. younger or nontechnical users). In particular, it should address the shortcomings from \secref{impact} around real-world impact. 
    \item \textbf{Effective legislation} of AI as it relates to this area is required. 
    There is emerging regulatory work, such as the EU AI act \cite{euaiact}, but it does not directly address loneliness or AI's effects on vulnerable users. 
    The risks outlined in our paper are not hypothetical: 
    consider the case of a US teen who committed suicide after allegedly being pushed to do so by a chatbot \cite{Payne2024}. 
    While the subsequent ruling that chatbots do not have free-speech rights, as the defence maintained \cite{Payne2025}, is a step in the right direction, this decision excludes loneliness as an explicit factor in the cause of death. 
    Legislation is not consistently global, but loneliness is; and so are its consequences. 
\end{enumerate}

\section{Limitations}\label{sec:limitations}
\subsection{Automated Annotation Reliability}
It is well-known that LLM annotators, such as GPT-4, exhibit biases and may not be reliable \cite{stureborg2024largelanguagemodelsinconsistent,doddapaneni2024findingblindspotsevaluator}, especially in multilingual scenarios \cite{hada-etal-2024-large,rtplx}. 
On the other hand, evidence exists of their usefulness in some scenarios \cite{zheng2023judging,chiang-lee-2023-large}. 
To address this ambiguity, we performed manual annotation and statistical analysis on top of the annotations. 
We found that the model is reliable but within a 1-5\% label-dependent margin of error. 

\subsection{Corpus Representativeness}
The corpus for WildChat was gathered via an API within Hugging Face. 
As pointed out by the authors, this might not be representative of the entire user base for ChatGPT. 
However, we believe it acts as a reasonable proxy given the volume of dialogues in the original corpus and the nature of the data we worked with. 

The representativeness of WildChat could also impact on the proportions of toxic and harmful dialogues: given that the access to the API was anonymous, there could be a higher-than-normal skewness towards this content. 
However, the numbers reported in the paper are well-below what we found (11\% versus 20\%). 
The disagreement in proportions does not affect our findings, as the focus of our work is different. 
Nonetheless, given our qualitative analysis of the toxic content, we still consider these types of interactions as concerning. 

\subsection{Loneliness Assessment}
The underlying assumption behind methods screening for conditions via text--including ours--is that people are comfortable enough to discuss their own concerns with the service. 
This assumption must hold: otherwise, scanning for loneliness would not be tenable. 
Our experimental process addressed this limitation by selecting and analysing lonely interactions by hand. 

\subsection{Impact of our Setup}\label{sec:impact}
Our setup allowed for both qualitative and quantitative analysis of a large volume of real, human-produced data. 
However, it has two downsides. First, it blurs out the distinction between the LLM behind the service and the `thing' (service, persona, etc) interacting with the user. 
This means that the conclusions we drew depend on the \textit{stack} (e.g., the UI, content moderators, etc), and not on the LLM alone. 
Holistically, this does not affect our work. However, in terms of actionable items for the NLP community, there must exist such a distinction. 
The second is that our approach relies solely on transcripts. This means that we can only draw conclusions based on what exists visible in the data, and impact outside of it may only be hypothesised. Nonetheless, it is worth noting that severe outcomes, such as loss of life, have occurred \cite{Payne2024,Walker2023}. 
Hence, we have extended recommendations specifically to address both limitations (\secref{recommendations}) and mitigate or eliminate real-world consequences. 

\section{Ethics}
Our work focused on the evaluation of a pre-existing, anonymised dataset. 
However, throughout the work we noticed that the anonymisation engine used typically failed for non-English names. 
Due to the sensitive nature of the data evaluated, along with licencing considerations, we only release the annotations and code to reproduce our analysis, but not the verbatim interactions. 
To discourage tracing, all interactions in this paper are paraphrased from the original data and the original dialogues are only available upon request.

\section*{Acknowledgments}
The author wishes to thank I. McCrum for comments on this work; and the anonymous reviewers, whose thorough feedback strengthened the arguments of this paper.

\DeclareRobustCommand{\DE}[3]{#2}
\DeclareRobustCommand{\VAN}[3]{#2}
\bibliography{biblio}

\appendix

\section{Prompts}\label{app:prompts}

The prompt we used to label the corpus are in \promptref{prompt1} (general labelling) and \promptref{prompt2} (loneliness analysis). 
The model we used is optimised to work with ChatML, a standard for model prompting, so the prompts in this section only represent the main instructions passed to the model and omit the exemplars and suggested labels. 
The reader is encouraged to review the code in the repository for full reproducibility. 
The prompts had reasonable accuracy, ranging from 86\% to 99\%; see \appref{reliability} for an analysis on this performance. 

\captionsetup[table]{name=Prompt}
\setcounter{table}{0}

\begin{table*}[th]
    \begin{tabular}{p{0.95\linewidth}}
    \toprule
\cellcolor{gray!5}You will be given a transcript between a chatbot and a human enclosed in <REQUESTS></REQUESTS> tags. \\
\cellcolor{gray!5}Each turn is enclosed in either <HU></HU> (for human) and <AI></AI> (for AI) \\
\cellcolor{gray!5}The transcript may be in English or any other language. \\
\cellcolor{gray!5}The transcript may be in English or any other language.\\
\cellcolor{gray!5}Your job will be to detect and label the intents from the *human* requests.\\
\cellcolor{gray!5}Intent = all comma-separated matching intents from {INTENTS}. Note that 'sexual content' includes fetish content such as diapers, scat, etc.\\
\cellcolor{gray!5}Note that the 'general conversation' intent is between the human and the AI. If the AI or human have dialogue as part of the input, it is \_not\_ general conversation.\\
\cellcolor{gray!5}'general conversation' and `inquiry' involve the user dialoguing with the AI, asking about what it thinks about a subject (`inquiry') or just maintaining a conversation ('general conversation').\\
\cellcolor{gray!5}If the Intent contains 'sexual content' or 'harmful content', also provide comma-separated Reasons and Target. \\
\cellcolor{gray!5}Available Reasons are {REASONS}. There can be more than one Reason. \\
\cellcolor{gray!5}You must also specify who is the Target of this harmful content: {TARGET}. \\
\cellcolor{gray!5}If `other', please specify it like so: other:label\\
\bottomrule
\end{tabular}
\caption{
System prompt used for labelling the data. 
The system prompt (above) and the exemplars (not pictured) are passed in to the model as a list of JSON entries. 
Although we pre-specified sets for Intent, Reasons and Targets (e.g., for the latter it was \{`men', `women', `minors', `other'\}) the model was encouraged to suggest labels that we later clustered manually. 
The Intent subset of this prompt had a 86 $\pm$ 4.7\% accuracy, and the reasons and targets had 99.2 $\pm$ 1.2\% accuracy, all at a 95\% CI.
}\label{pro:prompt1}
\end{table*}

\begin{table*}[th]
    \begin{tabular}{p{0.95\linewidth}}
    \toprule
\cellcolor{gray!5}You will be given a transcript between a chatbot and a human enclosed in <REQUESTS></REQUESTS> tags. \\
\cellcolor{gray!5}Each turn is enclosed in either <HU></HU> (for human) and <AI></AI> (for AI) \\
\cellcolor{gray!5}The transcript may be in English or any other language. \\
\cellcolor{gray!5}Your job will be to score the *human* dialogue based on the metrics below.\\
\cellcolor{gray!5}The metrics related to loneliness and how the writer in <HU></HU> is experiencing it.\\
\cellcolor{gray!5}Note: Loneliness is the lack of fulfilling relationships. Remember that one can be surrounded by people and still be lonely.\\
\cellcolor{gray!5}Start by determining whether the writer is lonely, and then fill out the metrics.\\
\cellcolor{gray!5}The metrics are:\\
\cellcolor{gray!5}Lonely: 0 or 1. 0 if not lonely, 1 if lonely.\\
\cellcolor{gray!5}Temporal: any of [`transient', `enduring', `ambiguous', 'N/A'].\\
\cellcolor{gray!5}Interaction: any of ['seeking advice', 'providing help', 'seeking validation and affirmation', 'reaching out', 'non directed interaction'].\\
\cellcolor{gray!5}Context: any of [`social', `physical', `somatic', `romantic', 'N/A'].\\
\cellcolor{gray!5}Interpersonal: any of [`romantic', `friendship', `family', `colleagues', `N/A'].\\
\cellcolor{gray!5}If it is not Lonely (Lonely=0), the values of Temporal, Interaction, Context, and Interpersonal are all N/A.\\
\cellcolor{gray!5}Otherwise, return them comma-separated.\\
\bottomrule
\end{tabular}
\caption{System prompt used for labelling the data in terms of loneliness, following the parameters from \citet{Jiang_Jiang_Leqi_Winkielman_2022}. 
We used the ChatML format: exemplars and the system prompt below are passed in to the model as a list of JSON entries. 
Due to the complex nature of this data, we solely used this prompt as a way to extract lonely interactions and did not perform a t-test, opting for the Reflexive Thematic Analysis instead.}\label{pro:prompt2}
\end{table*}

\section{Experimental Details}\label{app:calldetails}
We used GPT-4o (gpt4-o-2024-05-13) through the Azure OpenAI API. For our calls, we set the LLM temperature to zero and maximum return tokens to 128; and left the rest of parameters as default. 
All the data analysis was done in a consumer-grade laptop. 

\section{Labeller Reliability Analysis}\label{app:reliability}

To ensure the validity of our results we performed a student's t-test on a subset of the labelled corpus (n=250) to a 95\% CI, along with a qualitative analysis of the failed points. 
A t-test implicitly assumes a normal distribution for the underlying distribution. 
We consider this a reasonable assumption given the large size of WildChat. 
The accuracies per label were 86.4 $\pm$ 4.7\% for Intent, 99.2 $\pm$ 1.2\% for Reasons, and 99.2 $\pm$ 1.2\% for Target.
Overall, the model was able to recognise the specified intents to a reasonable accuracy, though our analysis showed that it sometimes skipped some acceptable labels (e.g., writing assistance with question answering) or confusing inquiries with question-answering. 
We attribute the high accuracy of Reasons and Target to the narrow label set used, as well as their low ambiguity.

\section{Corpus Breakdowns}\label{app:corpusbreakdown}
In this section we elaborate in the distinctions between the various subsets of WildChat used in our work. 
The \textit{main} corpus is the corpus sampled from WildChat, while the \textit{relevant} corpus is a subset of the main corpus that does not contain any task-oriented dialogue (i.e., only general conversation interactions). 
The \textit{lonely} dialogues are these dialogues that have been labelled as lonely. 
See \tabref{corpusbreakdown} for a breakdown of each of the subsets, along with volumes and descriptions. 

\captionsetup[table]{name=Table}
\setcounter{table}{2} %
\begin{table*}
    \centering
    \begin{tabular}{lll}
    \toprule
    \textbf{Subset} & \textbf{Description} & \textbf{Volume (interactions)} \\
    \midrule
    \textit{Main corpus} & Corpus subsampled from WildChat & 79,951 \\
    \textit{Relevant corpus} & Subset of the main corpus containing & \\ 
    & only general conversation & 30,481 \\ 
    \textit{Lonely corpus} & Subset of the main corpus of interactions &\\
    & labelled as lonely & 2,313; where 1,595 belong \\
    & &to the relevant corpus. \\ 
     \bottomrule
    \end{tabular}
    \caption{Naming for each of the subsets used in this paper, along with a description and the total number of interactions present in them.}
    \label{tab:corpusbreakdown}
\end{table*}

\section{Corpus Composition Analysis}\label{app:composition}

\begin{figure}
  \centering
  \includegraphics[width=\linewidth]{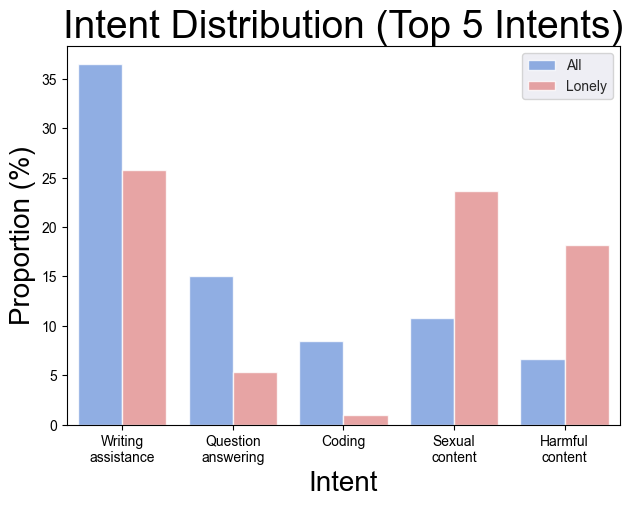} %
  \caption{Top five intents of the main corpus, compared with these of lonely users. There was a lower amount of writing assistance and coding intents, but sexual and harmful content is much higher: 7\% versus 18\% for harmful, and 11\% versus 24\% for sexual.}\label{fig:taxonomydistro}
\end{figure}

Prior to our analysis we clustered the LLM-suggested labels and added them to the taxonomy from \tabref{taxonomytable}. 
We show in \figref{taxonomydistro} the distribution for the top five intents over the main corpus, and the same distribution when ablated out by the interactions considered to be lonely. 
As noted, we observed a much higher incidence of toxic content in lonely interactions when compared to the main corpus (55\% versus 20\%). 
Namely, we observed 18\% versus 7\% for harmful content; and 24\% versus 11\% for sexual content.
While we are unable to explain why the model flagged this type of content as lonely, \promptref{prompt2} might offer a clue: the definition of loneliness includes unfulfilled relationships. 

\begin{figure}
  \centering
  \includegraphics[width=\linewidth]{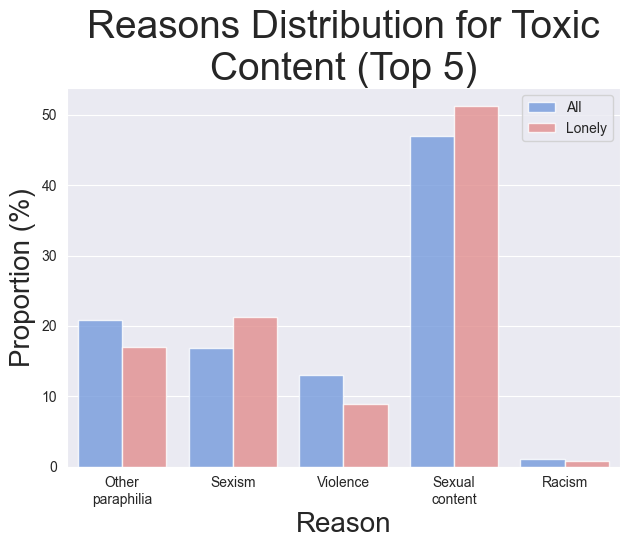} %
  \caption{Breakdown of the top five reasons for toxic content in our relevant corpus, compared with the subset of lonely users. 
  There are no considerable variations in this distribution, with perhaps a slightly higher (+4-6\%) proportion of sexism and sexual content.}\label{fig:resultsdistro}
\end{figure}

Further analysis of the reasons for selecting the toxic content (\figref{resultsdistro}) showed that there were frequent requests for paraphilia and other fetish content, thus supporting our hypothesis of unfulfilled relationships. 
Although we did not observe a significant change in the distribution of reasons, there was a slight (4\%) uptick in the amount of general sexual content and sexism, and slightly lower (-4\%) incidences of violence and fetish content. 

\begin{figure}
  \centering
  \includegraphics[width=\linewidth]{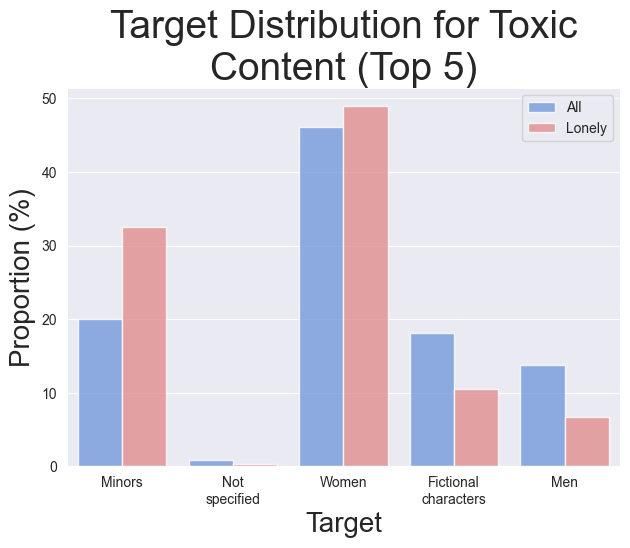} %
  \caption{Breakdown of the top five targets of toxic content in our corpus, compared with the subset for lonely users. 
  Most of the targets in the content remained steady between subsets, though there was a disproportionate (+12\%) amount of harmful content where a minor was a target. The frequency of toxic content where men are targets is half (14\% to 7\%) in lonely interactions. 
}\label{fig:targetdistro}
\end{figure}

We also show in \figref{targetdistro} the most common targets of toxic interactions. 
However, when looking at the distribution of targets for this toxic content, we noticed a disproportionately larger (+12\%) amount of content being directed at minors in the lonely interactions, coupled with a lower (-5\%) incidence of toxic content with men as the target.

\end{document}